\documentclass{article}
\pdfoutput=1

    \usepackage[final, nonatbib]{NeurIPS2020/neurips_2020}

\usepackage[utf8]{inputenc} 
\usepackage[T1]{fontenc}    
\usepackage{hyperref}       
\usepackage{url}            
\usepackage{booktabs}       
\usepackage{amsfonts}       
\usepackage{nicefrac}       
\usepackage{microtype}      

\usepackage{graphicx} 
\usepackage{amsfonts} 
\usepackage{amsmath,soul}
\usepackage{amssymb}
\usepackage[capitalize]{cleveref} 
\newcommand{\unit}[1]{\ensuremath{\, \mathrm{#1}}} 

\usepackage{dsfont} 
\usepackage[font=small,skip=0pt]{subcaption} 
\usepackage{balance}
\usepackage[font=small]{caption}
\usepackage{bm} 
\usepackage{float} 
\usepackage{wrapfig} 

\usepackage[authormarkuptext=name,addedmarkup=bf,authormarkupposition=left]{changes}
\definechangesauthor[name={X.~X.}, color={orange}]{xx}
\definechangesauthor[name={B.~L.}, color={MediumSeaGreen}]{bl}
\definechangesauthor[name={SRC}, color={brown}]{salva}

\title{Graph Neural Networks for Improved El Niño Forecasting}

\author{%
 Salva Rühling Cachay \\
 Technical University of Darmstadt \\
 \texttt{salvaruehling@gmail.com} \\
   \And
   Emma Erickson$^*$\\
   University of Illinois at Urbana-Champaign \\
   \And
   Arthur Fender C. Bucker$^*$ \\
   University of São Paulo \& TU Munich\\
     \And
   Ernest Pokropek$^*$ \\
   Warsaw University of Technology \\
    \And
   Willa Potosnak\thanks{Contributed equally as second authors.} \\
   Duquesne University \\

   \And
   Salomey Osei \\
   African Institute for Mathematical Sciences \\
     \And
   Bj\"orn L\"utjens \\
   Massachusetts Institute of Technology \\
}

\begin{document}

\maketitle
\begin{abstract}
Deep learning-based models have recently outperformed state-of-the-art seasonal forecasting models, such as for predicting El Ni\~no-Southern Oscillation (ENSO). However, current deep learning models are based on convolutional neural networks which are difficult to interpret and can fail to model large-scale atmospheric patterns called teleconnections. Hence, we propose the application of spatiotemporal Graph Neural Networks (GNN) to forecast ENSO at long lead times, finer granularity and improved predictive skill than current state-of-the-art methods.
The explicit modeling of information flow via edges may also allow for more interpretable forecasts.
Preliminary results are promising and outperform state-of-the art systems for projections 1 and 3 months ahead.
\end{abstract}


\vspace{-3mm}
\section{Introduction}\label{sec:intro}
\vspace{-5.6mm}
\begin{figure}[H]
 \centering
 \begin{subfigure}{1\columnwidth}
  \centering
      \includegraphics [trim=0 .3 0 0, clip, width=1., clip, width=1.\textwidth, angle = 0]{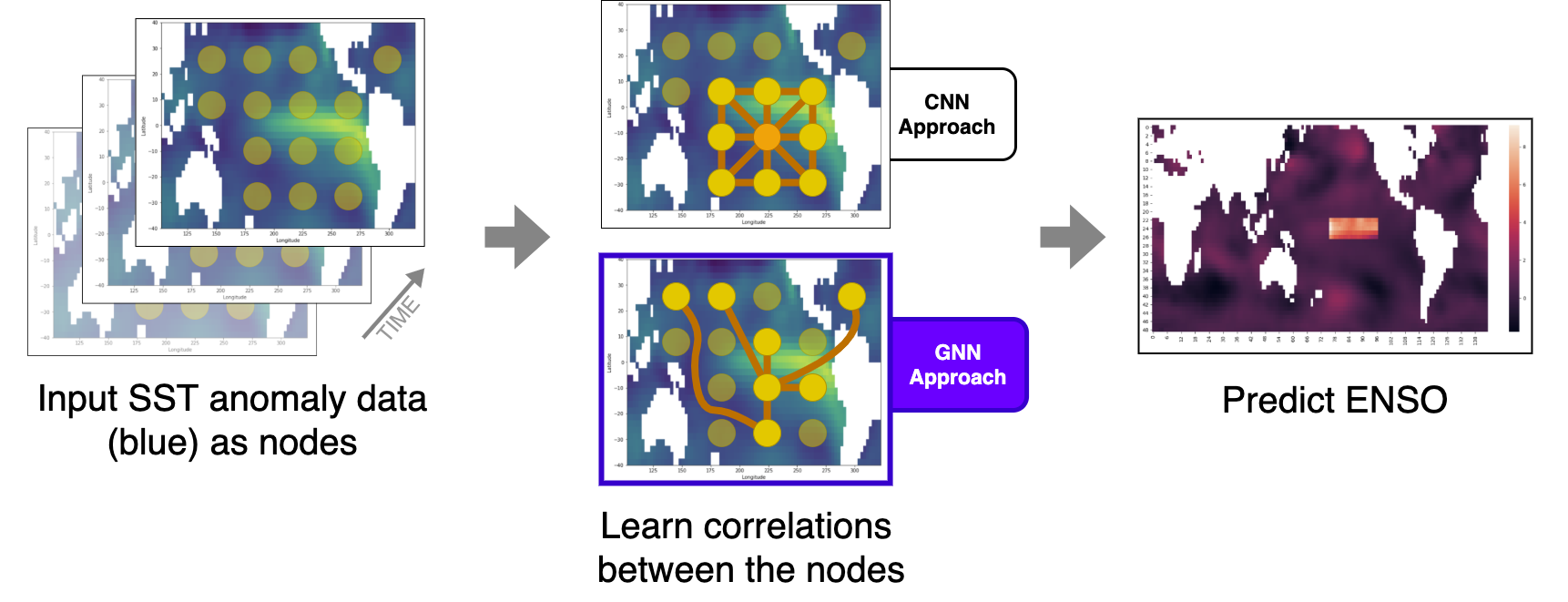}
      \vspace{.02in}
  \end{subfigure}
  \vspace*{-6mm}
\caption[teaserfigure]
{We propose spatiotemporal Graph Neural Networks (GNNs) to forecast ENSO.
GNNs can better exploit large-scale, spatiotemporal patterns indicative of ENSO than CNNs, which are based on local convolutions.
} 
\label{fig:teaser} 
\end{figure}
\vspace{-3mm}

El Niño–Southern Oscillation (ENSO) is an irregularly recurring phenomenon involving fluctuating temperatures---the alternation of warm El Niño and cold La Niña conditions---in the tropical Pacific Ocean. It is a major driver of climate variability, causes disasters such as floods, droughts and heavy rains in various regions of the world \cite{ENSO_impacts, ENSO_floods, ENSO_rainforest, ENSO_droughts, ENSO_hurricanes, ENSO_precipitation, ENSO_precipitation2} and has implications for agriculture \cite{ENSO_preds_for_agriculture, ENSO_agriculture, ENSO_agriculture2} and public health \cite{ENSO_health, ENSO_health_fire, ENSO_health2, ENSO_health_review}.
Worldwide teleconnections, i.e. interlinked, large-scale phenomena, as well as the high variability regarding its manifestations have kept long-term ENSO forecasts at traditionally low skill .

While previous studies indicate that more frequent, long-term or variable El Ni\~no conditions may result due to global warming from greenhouse gases~\cite{yeh2009nino, Higher_frequency_due_to_CC, rosenzweig_oscillation_impacts, EP_variable_due_to_CC}, the extent of influence climate change will have on ENSO is yet unknown and still debated given its complexity \cite{ENSO_and_CC, Uncertain_ClimateChange_predictions, enso_complexity_timmermann2018, ENSO_review, Cai2020ButterflyENSO}. This work proposes the first application of graph neural networks to seasonal forecasting and shows initial results that outperform existing dynamical and deep learning ENSO models for 1 and 3 lead months.

\section{Related Works}
\vspace{-3mm}
The forecasting methods in use can be broadly classified into dynamical and statistical systems \cite{ENSO_review, ENSO_predictability, ENS=_predictability2)}. The former are based on physical processes/climate models (e.g. atmosphere–ocean coupled models), while the latter are data-driven (including ML based approaches).  

\textbf{Machine Learning for ENSO forecasting} $\;$ Recently, deep learning was successfully used to forecast ENSO 1\unit{yr} ahead \cite{TemporalCNN_ENSO} as well as with a lead time of up to $1.5\unit{yrs}$~\cite{CNN_ENSO}, thus out-performing state-of-the-art dynamical methods. 
Both project the Oceanic Niño Index (ONI) for various lead times. The former only use the ONI index time series as input of a temporal Convolutional Neural Network (CNN), while the latter feed sea surface temperature (SST) and heat content anomaly maps data to a CNN. 
Most statistical methods can only predict the single-valued index, an averaged metric over SST anomalies that does not convey more zonal information. A notable exception, makes use of an encoder-decoder 
approach \cite{DLENSO}. An overview over other machine learning methods used to project ENSO, is given in \cite{ML_for_ENSO}.

\textbf{Climate networks} $\;$ In climate networks \cite{ClimateNetworks}, which stem from the field of complex networks, each grid cell of a climate dataset is considered as a network node and edges between pairs of nodes are set up using some similarity measure
. They have been used to detect and characterize SST teleconnections~\cite{agarwal2019network} and to successfully project ENSO $1\unit{yr}$ prior~\cite{ENSO_EarlyForecast_with_climateNetworks}. The latter exploits the observation that, a year before an ENSO event, a large-scale cooperative mode seems to link the equatorial Pacific corridor (“El Niño basin”) and the rest of the Pacific ocean.


\textbf{Graph neural networks} $\;$
In the past years, GNNs have surged as a popular sub-area of research within machine learning~\cite{GNN_review}. Interestingly, they have scarcely been used in earth and atmospheric sciences---a few applications using them for earthquake source detection \cite{van2020Seisms_and_GNNs}, power outage prediction \cite{owerko2018Outages_and_GNNs} and wind-farm power estimation \cite{park2019Wind_power_and_GNNs}. GNNs have just recently been extended to spatiotemporal settings, with a focus on traffic forecasting~\cite{MTGNN, Syncronous_STGCN, WaveNet, STGCN}. 

Our GNN approach to ENSO forecasting builds on the climate network’s precedent of describing climate as a network of nodes related by
non-local connections. Based on this precedent and the recent success of GNNs for spatiotemporal tasks, it is expected that spatiotemporal
GNNs will be able to learn the large-scale dependencies in between climate nodes and accurately model the inherent complexity of the ENSO
phenomenon. We are currently extending a state-of-the-art spatiotemporal GNN architecture \cite{MTGNN}, that does not require pre-defined edges and supports multi-step forecasting, to the domain of ENSO forecasting.


\section{Data}\label{sec:data}
\vspace{-3mm}
ENSO depends on and affects different environmental factors. Amongst these are sea-level pressure, zonal and meridional components of the surface wind, sea surface temperature, surface air temperature \cite{enso_complexity_timmermann2018}.
The climate variable, time series datasets of interest for this research are: \begin{itemize}
    \item 
    NOAA ERSSTv5 \cite{ERSSTv5},  with SST data recorded since 1854, that we have used for our preliminary experiments (we train on 1871-1973 and test on the 1984-2020 period)
    \item
    Coupled Model Intercomparison Project phase5 (CMIP5) \cite{CMIP5} historical simulations recorded since 1861, that are particularly interesting for pre-training the model since only few observational data are available
    \item
    Simple Ocean Data Assimilation (SODA) \cite{SODA}
    , reanalysis data recorded from 1870-2010
    \item
    Global Ocean Data Assimilation System (GODAS) \cite{GODAS}
    reanalysis data (since 1980).
\end{itemize} 
The last three datasets are open-source in the processed form they were used by \cite{CNN_ENSO} for pre-training, training and testing, respectively. 
The suitability of these datasets to deep learning methods has been demonstrated by \cite{CNN_ENSO}. Preliminary analysis will focus on these datasets, but more datasets may be incorporated to include other relevant variables, such as sea-level pressure and surface wind.

\section{Proposed model}\label{sec:model_architecture}
\vspace{-2.5mm}
Graph Neural Networks (GNN) generalize the notion of locality that is exploited by Convolutional Neural Networks (CNN), allowing us to model arbitrarily complex connections that are paramount for long-term forecasts of phenomena like ENSO, where relations are non-Euclidean.
Importantly, CNNs assume translation equivariance of the input \cite{DL_book}. For seasonal forecasts, however,  spatially shared representations for the globe do not seem adequate, since it does matter where exactly a certain phenomenon or pattern occurs (e.g. at a teleconnection versus at a distant, unrelated part of the world). Additionally, GNNs are more efficient than recurrent neural networks and LSTMs \cite{STGCN}, which are often used in ENSO forecasting models \cite{DLENSO, climateNetworks_and_LSTM}.
 
Climate datasets are often gridded, therefore, the grid cells (i.e. geographical locations) can be naturally mapped to the nodes of a GNN. The graph's edges, which model the flow of information between nodes, are the main argument in favor of a GNN approach. Edges can be chosen based on mid- and long-range climate dependencies (e.g. based on domain expertise or on edges analyzed in climate networks research), or they can be inferred by the GNN using recent graph structure learning approaches \cite{MTGNN}. The explicit modeling of interdependencies based on domain expertise, or the GNN's choice of meaningful edges (e.g. well known patterns or teleconnections), greatly enhances the model's interpretability.

Moreover, most statistical methods only forecast the single-valued index and not the zonal sea surface temperature (SST) anomalies (which can be used for, e.g., ENSO type classification \cite{ENSO_review} and a more informed forecast).
A GNN can naturally overcome these limitations by forecasting the target variable at the nodes---which correspond to geographical regions---of interest (in our case the SST anomalies in the ONI region).
The multiple spatiotemporal GNN architectures that have been recently proposed seem particularly well suited as a starting point \cite{MTGNN, Syncronous_STGCN, WaveNet, STGCN}.
A high-level visualization of our approach is illustrated in~\cref{fig:teaser}.

\section{Preliminary results}\label{sec:results}
\vspace{-3mm}
The presence of an ENSO event is commonly measured via the running mean over $k$ months of sea surface temperature anomalies (SSTA) over the Oceanic Niño Index (ONI, $k=3$) region (5N-5S, 120-170W), also known as the Niño3.4 index region ($k=5$).

\vspace{-1mm}
\begin{wraptable}{r}{6.5cm}
\vspace{-3.5mm}
\caption{Correlation skill for $n$ lead months}
    \label{table:results}
    \centering
    \begin{tabular}{@{} *4l @{}}
    \toprule
     Model     & $n=1$ & $n=3$ & $n=6$\\ 
    \midrule
    CNN~\cite{CNN_ENSO} & $\approx$ 0.94 & 0.8742 & 0.7616 \\
    GNN (ours)          & 0.9867 & 0.8936 & 0.6776 \\
    \bottomrule
    \end{tabular}
    \vspace{-3.1mm}
\end{wraptable}
Preliminary work using SSTAs computed from the ERSSTv5 dataset as input to the spatiotemporal GNN proposed in \cite{MTGNN}, shows promising results in predicting the ONI index for up to 6\unit{mon} ahead forecasts (\cref{table:results}). \\ 
We use the SST anomalies within the ONI region over 3\unit{mon}, and a simple
architecture with only two layers and no pre-defined edges.
Longer lead times were not yet satisfying, which we expect to be caused by 1) the small dataset (1233 data points in the training set), which we hope to overcome by using transfer learning like \cite{CNN_ENSO}; 2) while SST anomalies are good short-term predictors of ENSO, long-term ENSO projections usually rely on other variables such as heat content anomalies, which we aim to incorporate in our model.

\section{Discussion and Future Works}
\vspace{-2.5mm}

An improved model could have a significant impact on global seasonal climate prediction, due to ENSOs teleconnections.  Leveraged as a tool by climate researchers, longer lead-time predictions would provide more time to determine the potential impact of the phenomenon. These lead-times would allow those in the various impacted areas to prepare for and adapt to the predicted climate and its effects on industry, agriculture, safety, and human quality of life.

In addition to helping populations impacted by ENSO, a successful deployment of a GNN architecture for ENSO forecasting would show its suitability to non-linear and complex earth and atmospheric modeling in general, such as projection of other oscillations or weather forecasting.

Finally, future work might explore using ENSO indicators as predictors in the GNN, forecasting ENSO's impacts (such as precipitation) across the globe due to teleconnections.

\clearpage
\subsubsection*{Acknowledgments}
We would like to thank the ProjectX organizing committee for motivating this work. We gratefully acknowledge the computational support by the Microsoft AI for Earth Grant. We would also like to thank Captain John Radovan for sharing his expertise regarding the current ENSO models and their global applications, Suyash Bire for his guidance on ENSO model limitations and result interpretation, as well as Chen Wang for his guidance on GNN architecture.

Björn Lütjens' research has been sponsored by the United States Air Force Research Laboratory and the United States Air Force Artificial Intelligence Accelerator and was accomplished under Cooperative Agreement Number FA8750-19-2-1000. The views and conclusions contained in this document are those of the authors and should not be interpreted as representing the official policies, either expressed or implied, of the United States Air Force or the U.S. Government. The U.S. Government is authorized to reproduce and distribute reprints for Government purposes notwithstanding any copyright notation herein.

\bibliographystyle{IEEEtran}
\bibliography{references}

\begin{thebibliography}{10}
\providecommand{\url}[1]{#1}
\csname url@samestyle\endcsname
\providecommand{\newblock}{\relax}
\providecommand{\bibinfo}[2]{#2}
\providecommand{\BIBentrySTDinterwordspacing}{\spaceskip=0pt\relax}
\providecommand{\BIBentryALTinterwordstretchfactor}{4}
\providecommand{\BIBentryALTinterwordspacing}{\spaceskip=\fontdimen2\font plus
\BIBentryALTinterwordstretchfactor\fontdimen3\font minus
  \fontdimen4\font\relax}
\providecommand{\BIBforeignlanguage}[2]{{%
\expandafter\ifx\csname l@#1\endcsname\relax
\typeout{** WARNING: IEEEtran.bst: No hyphenation pattern has been}%
\typeout{** loaded for the language `#1'. Using the pattern for}%
\typeout{** the default language instead.}%
\else
\language=\csname l@#1\endcsname
\fi
#2}}
\providecommand{\BIBdecl}{\relax}
\BIBdecl

\bibitem{ENSO_impacts}
J.~Lin and T.~Qian, ``A new picture of the global impacts of el nino-southern
  oscillation,'' \emph{Scientific Reports}, vol.~9, 12 2019.

\bibitem{ENSO_floods}
P.~J. Ward, B.~Jongman, M.~Kummu, M.~D. Dettinger, F.~C.~S. Weiland, and H.~C.
  Winsemius, ``Strong influence of el ni{\~n}o southern oscillation on flood
  risk around the world,'' \emph{Proceedings of the National Academy of
  Sciences}, vol. 111, no.~44, pp. 15\,659--15\,664, 2014.

\bibitem{ENSO_rainforest}
G.~B. Williamson, W.~F. Laurance, A.~A. Oliveira, P.~Delam{\^o}nica, C.~Gascon,
  T.~E. Lovejoy, and L.~Pohl, ``Amazonian tree mortality during the 1997 el
  nino drought,'' \emph{Conservation Biology}, vol.~14, no.~5, pp. 1538--1542,
  2000.

\bibitem{ENSO_droughts}
F.~Siegert, G.~Ruecker, A.~Hinrichs, and A.~Hoffmann, ``Increased damage from
  fires in logged forests during droughts caused by el nino,'' \emph{Nature},
  vol. 414, no. 6862, pp. 437--440, 2001.

\bibitem{ENSO_hurricanes}
J.~P. Donnelly and J.~D. Woodruff, ``Intense hurricane activity over the past
  5,000 years controlled by el ni{\~n}o and the west african monsoon,''
  \emph{Nature}, vol. 447, no. 7143, pp. 465--468, 2007.

\bibitem{ENSO_precipitation}
C.~F. Ropelewski and M.~S. Halpert, ``Global and regional scale precipitation
  patterns associated with the el ni{\~n}o/southern oscillation,''
  \emph{Monthly weather review}, vol. 115, no.~8, pp. 1606--1626, 1987.

\bibitem{ENSO_precipitation2}
------, ``North american precipitation and temperature patterns associated with
  the el ni{\~n}o/southern oscillation (enso),'' \emph{Monthly Weather Review},
  vol. 114, no.~12, pp. 2352--2362, 1986.

\bibitem{ENSO_preds_for_agriculture}
A.~R. Solow, R.~F. Adams, K.~J. Bryant, D.~M. Legler, J.~J. O'brien, B.~A.
  McCarl, W.~Nayda, and R.~Weiher, ``The value of improved enso prediction to
  us agriculture,'' \emph{Climatic change}, vol.~39, no.~1, pp. 47--60, 1998.

\bibitem{ENSO_agriculture}
\BIBentryALTinterwordspacing
R.~M. Adams, C.-C. Chen, B.~A. McCarl, and R.~F. Weiher, ``The economic
  consequences of enso events for agriculture,'' \emph{Climate Research},
  vol.~13, no.~3, pp. 165--172, 1999. [Online]. Available:
  \url{http://www.jstor.org/stable/24866033}
\BIBentrySTDinterwordspacing

\bibitem{ENSO_agriculture2}
W.~Anderson, R.~Seager, W.~Baethgen, and M.~Cane, ``Trans-pacific enso
  teleconnections pose a correlated risk to agriculture,'' \emph{Agricultural
  and forest meteorology}, vol. 262, pp. 298--309, 2018.

\bibitem{ENSO_health}
R.~S. Kovats, M.~J. Bouma, S.~Hajat, E.~Worrall, and A.~Haines, ``El ni{\~n}o
  and health,'' \emph{The Lancet}, vol. 362, no. 9394, pp. 1481--1489, 2003.

\bibitem{ENSO_health_fire}
M.~E. Marlier, R.~S. DeFries, A.~Voulgarakis, P.~L. Kinney, J.~T. Randerson,
  D.~T. Shindell, Y.~Chen, and G.~Faluvegi, ``El ni{\~n}o and health risks from
  landscape fire emissions in southeast asia,'' \emph{Nature climate change},
  vol.~3, no.~2, pp. 131--136, 2013.

\bibitem{ENSO_health2}
J.~A. Patz, D.~Campbell-Lendrum, T.~Holloway, and J.~A. Foley, ``Impact of
  regional climate change on human health,'' \emph{Nature}, vol. 438, no. 7066,
  pp. 310--317, 2005.

\bibitem{ENSO_health_review}
G.~R. McGregor and K.~Ebi, ``El ni{\~n}o southern oscillation (enso) and
  health: an overview for climate and health researchers,'' \emph{Atmosphere},
  vol.~9, no.~7, p. 282, 2018.

\bibitem{yeh2009nino}
S.-W. Yeh, J.-S. Kug, B.~Dewitte, M.-H. Kwon, B.~P. Kirtman, and F.-F. Jin,
  ``El ni{\~n}o in a changing climate,'' \emph{Nature}, vol. 461, no. 7263, pp.
  511--514, 2009.

\bibitem{Higher_frequency_due_to_CC}
W.~Cai, S.~Borlace, M.~Lengaigne, P.~Van~Rensch, M.~Collins, G.~Vecchi,
  A.~Timmermann, A.~Santoso, M.~J. McPhaden, L.~Wu \emph{et~al.}, ``Increasing
  frequency of extreme el ni{\~n}o events due to greenhouse warming,''
  \emph{Nature climate change}, vol.~4, no.~2, pp. 111--116, 2014.

\bibitem{rosenzweig_oscillation_impacts}
C.~Rosenzweig and D.~Hillel, \emph{Climate Variability and the Global Harvest:
  Impacts of El Niño and Other Oscillations on Agro-Ecosystems}.\hskip 1em
  plus 0.5em minus 0.4em\relax New York, N.Y.: Oxford University Press, 2008.

\bibitem{EP_variable_due_to_CC}
W.~Cai, G.~Wang, B.~Dewitte, L.~Wu, A.~Santoso, K.~Takahashi, Y.~Yang,
  A.~Carr{\'e}ric, and M.~J. McPhaden, ``Increased variability of eastern
  pacific el ni{\~n}o under greenhouse warming,'' \emph{Nature}, vol. 564, no.
  7735, pp. 201--206, 2018.

\bibitem{ENSO_and_CC}
W.~Cai, A.~Santoso, G.~Wang, S.-W. Yeh, S.-I. An, K.~M. Cobb, M.~Collins,
  E.~Guilyardi, F.-F. Jin, J.-S. Kug \emph{et~al.}, ``Enso and greenhouse
  warming,'' \emph{Nature Climate Change}, vol.~5, no.~9, pp. 849--859, 2015.

\bibitem{Uncertain_ClimateChange_predictions}
\BIBentryALTinterwordspacing
H.~Paeth, A.~Scholten, P.~Friederichs, and A.~Hense, ``Uncertainties in climate
  change prediction: El niño-southern oscillation and monsoons,'' \emph{Global
  and Planetary Change}, vol.~60, no.~3, pp. 265 -- 288, 2008. [Online].
  Available:
  \url{http://www.sciencedirect.com/science/article/pii/S092181810700046X}
\BIBentrySTDinterwordspacing

\bibitem{enso_complexity_timmermann2018}
A.~Timmermann, S.-I. An, J.-S. Kug, F.-F. Jin, W.~Cai, A.~Capotondi, K.~M.
  Cobb, M.~Lengaigne, M.~J. McPhaden, M.~F. Stuecker \emph{et~al.}, ``El
  ni{\~n}o--southern oscillation complexity,'' \emph{Nature}, vol. 559, no.
  7715, pp. 535--545, 2018.

\bibitem{ENSO_review}
\BIBentryALTinterwordspacing
C.~Wang, C.~Deser, J.-Y. Yu, P.~DiNezio, and A.~Clement, \emph{El Ni{\~{n}}o
  and Southern Oscillation (ENSO): A Review}.\hskip 1em plus 0.5em minus
  0.4em\relax Dordrecht: Springer Netherlands, 2017, pp. 85--106. [Online].
  Available: \url{https://doi.org/10.1007/978-94-017-7499-4_4}
\BIBentrySTDinterwordspacing

\bibitem{Cai2020ButterflyENSO}
W.~Cai, B.~Ng, T.~Geng, L.~Wu, A.~Santoso, and M.~McPhaden, ``Butterfly effect
  and a self-modulating el ni{\~n}o response to global warming.''
  \emph{Nature}, vol. 585 7823, pp. 68--73, 2020.

\bibitem{ENSO_predictability}
\BIBentryALTinterwordspacing
D.~Chen, M.~A. Cane, A.~Kaplan, S.~E. Zebiak, and D.~Huang, ``Predictability of
  el niño over the past 148 years,'' \emph{Nature}, vol. 428, no. 6984, p.
  733—736, April 2004. [Online]. Available:
  \url{https://doi.org/10.1038/nature02439}
\BIBentrySTDinterwordspacing

\bibitem{ENS=_predictability2)}
\BIBentryALTinterwordspacing
A.~V. Fedorov, S.~L. Harper, S.~G. Philander, B.~Winter, and A.~Wittenberg,
  ``{How Predictable is El Niño?}'' \emph{Bulletin of the American
  Meteorological Society}, vol.~84, no.~7, pp. 911--920, 07 2003. [Online].
  Available: \url{https://doi.org/10.1175/BAMS-84-7-911}
\BIBentrySTDinterwordspacing

\bibitem{TemporalCNN_ENSO}
J.~Yan, L.~Mu, L.~Wang, R.~Ranjan, and A.~Y. Zomaya, ``temporal convolutional
  networks for the advance prediction of enso,'' \emph{Scientific Reports},
  vol.~10, no.~1, pp. 1--15, 2020.

\bibitem{CNN_ENSO}
Y.-G. Ham, J.-H. Kim, and J.-J. Luo, ``Deep learning for multi-year enso
  forecasts,'' \emph{Nature}, vol. 573, pp. 568--572, 9 2019.

\bibitem{DLENSO}
D.~He, P.~Lin, H.~Liu, L.~Ding, and J.~Jiang, ``Dlenso: A deep learning enso
  forecasting model,'' in \emph{PRICAI}, 2019.

\bibitem{ML_for_ENSO}
H.~Dijkstra, E.~Hernandez-Garcia, C.~Lopez \emph{et~al.}, ``The application of
  machine learning techniques to improve el nino prediction skill,''
  \emph{Frontiers in Physics}, vol.~7, p. 153, 2019.

\bibitem{ClimateNetworks}
\BIBentryALTinterwordspacing
A.~A. Tsonis, K.~L. Swanson, and P.~J. Roebber, ``{What Do Networks Have to Do
  with Climate?}'' \emph{Bulletin of the American Meteorological Society},
  vol.~87, no.~5, pp. 585--596, 05 2006. [Online]. Available:
  \url{https://doi.org/10.1175/BAMS-87-5-585}
\BIBentrySTDinterwordspacing

\bibitem{agarwal2019network}
A.~Agarwal, L.~Caesar, N.~Marwan, R.~Maheswaran, B.~Merz, and J.~Kurths,
  ``Network-based identification and characterization of teleconnections on
  different scales,'' \emph{Scientific Reports}, vol.~9, no.~1, pp. 1--12,
  2019.

\bibitem{ENSO_EarlyForecast_with_climateNetworks}
\BIBentryALTinterwordspacing
J.~Ludescher, A.~Gozolchiani, M.~I. Bogachev, A.~Bunde, S.~Havlin, and H.~J.
  Schellnhuber, ``Very early warning of next el ni{\~n}o,'' \emph{Proceedings
  of the National Academy of Sciences}, vol. 111, no.~6, pp. 2064--2066, 2014.
  [Online]. Available: \url{https://www.pnas.org/content/111/6/2064}
\BIBentrySTDinterwordspacing

\bibitem{GNN_review}
Z.~Wu, S.~Pan, F.~Chen, G.~Long, C.~Zhang, and S.~Y. Philip, ``A comprehensive
  survey on graph neural networks,'' \emph{IEEE Transactions on Neural Networks
  and Learning Systems}, 2020.

\bibitem{van2020Seisms_and_GNNs}
M.~P. van~den Ende and J.-P. Ampuero, ``Automated seismic source
  characterisation using deep graph neural networks,'' \emph{Geophysical
  Research Letters}, p. e2020GL088690, 2020.

\bibitem{owerko2018Outages_and_GNNs}
D.~Owerko, F.~Gama, and A.~Ribeiro, ``Predicting power outages using graph
  neural networks,'' in \emph{2018 IEEE Global Conference on Signal and
  Information Processing (GlobalSIP)}.\hskip 1em plus 0.5em minus 0.4em\relax
  IEEE, 2018, pp. 743--747.

\bibitem{park2019Wind_power_and_GNNs}
J.~Park and J.~Park, ``Physics-induced graph neural network: An application to
  wind-farm power estimation,'' \emph{Energy}, vol. 187, p. 115883, 2019.

\bibitem{MTGNN}
\BIBentryALTinterwordspacing
Z.~Wu, S.~Pan, G.~Long, J.~Jiang, X.~Chang, and C.~Zhang, ``Connecting the
  dots: Multivariate time series forecasting with graph neural networks,'' in
  \emph{Proceedings of the 26th ACM SIGKDD International Conference on
  Knowledge Discovery and Data Mining}, ser. KDD '20.\hskip 1em plus 0.5em
  minus 0.4em\relax New York, NY, USA: Association for Computing Machinery,
  2020, p. 753–763. [Online]. Available:
  \url{https://doi.org/10.1145/3394486.3403118}
\BIBentrySTDinterwordspacing

\bibitem{Syncronous_STGCN}
C.~Song, Y.~Lin, S.~Guo, and H.~Wan, ``Spatial-temporal synchronous graph
  convolutional networks: A new framework for spatial-temporal network data
  forecasting,'' in \emph{Proceedings of the AAAI Conference on Artificial
  Intelligence}, vol.~34, no.~01, 2020, pp. 914--921.

\bibitem{WaveNet}
\BIBentryALTinterwordspacing
Z.~Wu, S.~Pan, G.~Long, J.~Jiang, and C.~Zhang,
  ``\BIBforeignlanguage{English}{Graph wavenet for deep spatial-temporal graph
  modeling},'' in \emph{\BIBforeignlanguage{English}{Proceedings of the
  Twenty-Eighth International Joint Conference on Artificial Intelligence}},
  S.~Kraus, Ed., 2019, pp. 1907--1913, international Joint Conference on
  Artificial Intelligence 2019, IJCAI-19. [Online]. Available:
  \url{https://ijcai19.org/}
\BIBentrySTDinterwordspacing

\bibitem{STGCN}
\BIBentryALTinterwordspacing
B.~Yu, H.~Yin, and Z.~Zhu, ``Spatio-temporal graph convolutional networks: A
  deep learning framework for traffic forecasting,'' in \emph{Proceedings of
  the Twenty-Seventh International Joint Conference on Artificial Intelligence,
  {IJCAI-18}}, 7 2018, pp. 3634--3640. [Online]. Available:
  \url{https://doi.org/10.24963/ijcai.2018/505}
\BIBentrySTDinterwordspacing

\bibitem{ERSSTv5}
\BIBentryALTinterwordspacing
B.~Huang, P.~W. Thorne, V.~F. Banzon, T.~Boyer, G.~Chepurin, J.~H. Lawrimore,
  M.~J. Menne, T.~M. Smith, R.~S. Vose, and H.-M. Zhang, ``{Extended
  Reconstructed Sea Surface Temperature, Version 5 (ERSSTv5): Upgrades,
  Validations, and Intercomparisons},'' \emph{Journal of Climate}, vol.~30,
  no.~20, pp. 8179--8205, 09 2017. [Online]. Available:
  \url{https://doi.org/10.1175/JCLI-D-16-0836.1}
\BIBentrySTDinterwordspacing

\bibitem{CMIP5}
K.~E. Taylor, R.~J. Stouffer, and G.~A. Meehl, ``An overview of cmip5 and the
  experiment design,'' \emph{Bulletin of the American Meteorological Society},
  vol.~93, no.~4, pp. 485--498, 2012.

\bibitem{SODA}
\BIBentryALTinterwordspacing
B.~S. Giese and S.~Ray, ``El niño variability in simple ocean data
  assimilation (soda), 1871–2008,'' \emph{Journal of Geophysical Research:
  Oceans}, vol. 116, no.~C2, 2011. [Online]. Available:
  \url{https://agupubs.onlinelibrary.wiley.com/doi/abs/10.1029/2010JC006695}
\BIBentrySTDinterwordspacing

\bibitem{GODAS}
D.~Behringer and Y.~Xue, ``Evaluation of the global ocean data assimilation
  system at ncep: The pacific ocean,'' 2004.

\bibitem{DL_book}
I.~Goodfellow, Y.~Bengio, and A.~Courville, \emph{Deep learning}, 2016.

\bibitem{climateNetworks_and_LSTM}
C.~Broni-Bedaiko, F.~A. Katsriku, T.~Unemi, M.~Atsumi, J.-D. Abdulai,
  N.~Shinomiya, and E.~Owusu, ``El ni{\~n}o-southern oscillation forecasting
  using complex networks analysis of lstm neural networks,'' \emph{Artificial
  Life and Robotics}, vol.~24, no.~4, pp. 445--451, 2019.

\end{thebibliography}

\newpage




\end{document}